\documentclass{article}
\pdfoutput=1

     \usepackage[nonatbib]{neurips_2021}
 \usepackage{graphicx}
 \usepackage[ruled,vlined]{algorithm2e}
 \usepackage{subfig}
\usepackage[utf8]{inputenc} 
\usepackage[T1]{fontenc}    
\usepackage{hyperref}       
\usepackage{url}            
\usepackage{booktabs}       
\usepackage{amsfonts}       
\usepackage{nicefrac}       
\usepackage{microtype}      
\usepackage{xcolor}         
\usepackage[
backend=biber,
style=numeric,
sorting=ynt
]{biblatex}

\title{About Explicit Variance Minimization: Training Neural Networks for Medical Imaging  With Limited Data Annotations.}

\begin{document}
\nolinenumbers
\maketitle

\begin{abstract}
Self-supervised learning methods for computer vision have demonstrated the effectiveness of pre-training feature representations, resulting in well-generalizing Deep Neural Networks, even if the annotated data are limited. However, representation learning techniques require a significant amount of time for model training, with most of the time spent on precise hyper-parameter optimization and selection of augmentation techniques. We hypothesized that if the annotated dataset has enough morphological diversity to capture the diversity of the general population, as is common in medical imaging due to conserved similarities of tissue morphology, the variance error of the trained model is the dominant component of the Bias-Variance Trade-off. Therefore, we proposed the Variance Aware Training (VAT) method that exploits this data property by introducing the variance error into the model loss function, thereby, \textit{explicitly} regularizing the model. Additionally, we provided a theoretical formulation and proof of the proposed method to aid  interpreting the approach. Our method requires selecting only one hyper-parameter and matching or improving the performance of state-of-the-art self-supervised methods while achieving an order of magnitude reduction in the GPU training time. We validated VAT on three medical imaging datasets from diverse domains and for various learning objectives. These included a Magnetic Resonance Imaging (MRI) dataset for the heart semantic segmentation (MICCAI 2017 ACDC challenge), fundus photography dataset for ordinary regression of diabetic retinopathy progression (Kaggle 2019 APTOS Blindness Detection challenge), and classification of histopathologic scans of lymph node sections (PatchCamelyon dataset). Our code is available at \textit{https://github.com/DmitriiShubin/Variance-Aware-Training}.
\end{abstract}

\section{Introduction}
Limited data annotations is a mainstream problem for medical imaging due to the high cost of the annotation procedure, which creates challenges in cross-subject generalization. Various domain adaptation methods have been proposed to address this issue, such as Semi-Supervised (SeSL) methods \cite{Feng2020DMTDM,Bai2017SemisupervisedLF,Wang2020FocalMixSL} and a broad set of Self-Supervised (SSL) methods based on learning semantic similarity \cite{hinton,dos_2014}, context understanding \cite{ssl_puzzle,ssl_puzzle3,ssl_puzzle2,haghighi2021transferable,ssl_position,Gidaris2018UnsupervisedRL,Doersch2015UnsupervisedVR}, and medical imaging domain properties \cite{Jamaludin2017SelfsupervisedLF,Zheng20183DCA,Ouyang2020SelfSupervisionWS,Srinidhi2021SelfsupervisedDC,Chaitanya2020ContrastiveLO}.

While SeSL methods are generally time-consuming, requiring backpropagation through the entire dataset over multiple iterations, SSL methods are sensitive to the selection of correct augmentation transformations, making it time-consuming with respect to hyper-parameter optimization.

Our motivation stems from the field of machine learning in healthcare, and specifically medical imaging. We identified two main attributes of the data in the medical imaging domain: (1) the limited amount of well-labelled data, and (2) the often relatively low inter-subject and inter-subset variability compared to the entire population distribution. For example, the image acquisition protocols are conserved across institutions, there are limited vendors and types of devices, and for a given imaging interpretation task, there are a limited number of underlying pathological processes. This skewed partition of the overall variance suggests that a relatively small population sample might capture most of the total variance, and a learning algorithm that can be tuned according to this property might still be as effective while requiring much less compute resources than a more brute force approach. 

Following this assumption, we hypothesize that the limited dataset size causes the variance error of the trained model to be the dominant component in Bias-Variance Trade-off Decomposition (BVTD). A common approach to reducing the variance error of a Deep Neural Network (DNN) is to introduce regularisation into the model, such as dropout \cite{Srivastava2014DropoutAS} or weight decay. However, the major disadvantage of such methods is they reduce the explicit power of the model. In contrast, we propose a method that explicitly optimizes the variance, where the variance of the model is minimized directly as a part of the model's cost function, which introduces this constraint into the model without the loss of capacity.

Our work presents three major contributions. In Section 3.1, we propose a theorem describing the model's variance error behaviour and propose the objective function that enables explicit variance error optimization. By adopting and combining ideas from Generative Adversarial \cite{Goodfellow2014GenerativeAN}, Domain Adversarial \cite{uada}, and Siamese \cite{Bromley1993SignatureVU} Neural Networks, in Section 3.2 we propose the Variance-Aware Training (VAT) method, which is a time-efficient method for training generalized Convolution Neural Networks (CNN). Lastly, In Section 4, we provided benchmarks to quantify the performance of the proposed method and provided an interpretation based on the theoretical background presented in Section 3.1.

\section{Related work}
Prior theoretical work provided by Neal et al. \cite{Neal2018AMT} and Yang et el. \cite{Yang2020RethinkingBT} analyzed techniques for measuring the Bias-Variance Trade-off components of Deep Neural Networks. In this paper, we provide an alternative perspective to the BVTD, which is to consider it as the decomposition of the universal approximate error. Additionally, we have translated the described theoretical background into the novel variance-aware loss function, augmenting the arbitrary task-specific loss, that can be used to minimize the variance directly.

Fundamentally, our work is based on the principles described by Goodfellow et al. (2014) \cite{Goodfellow2014GenerativeAN} for Generative Adversarial Networks (GAN), where the mechanism minimizes the nonconformity between the distribution of real images and images generated by the generative neural network $G$ via backpropagation of the error from the discriminative network $D$.

In particular, our method is a special case of Domain-Adversarial Networks (DAN), which were introduced by Ganin et al. (2016) \cite{uada} and extended in applications by Kim et al. \cite{kim_2019}. DAN is designed to minimize a gap between two datasets from various domains using a Reversed Gradient Layer (RGL) to reverse all gradients passed through it. In our method, we used RGL to minimize a gap between training and validation subsets, thus minimizing the model's variance error component. Additionally, unlike the original DAN, we introduced our own computationally efficient feature map estimates to penalize both local and global feature representations.

As a feature extractor for our training method, we used the Siamese Neural Network (SNN) architecture proposed by Bromley et al. (1994) \cite{Bromley1993SignatureVU}. SNN was designed for learning feature representations as a similarity between a pair of samples through feature extractors with shared weights. Several later studies proposed an extension of the method to imaging data \cite{FeiFei2003ABA}, and afterward, diverse loss functions \cite{siamese_contrastive,Ghojogh,Hoffer} for applications such as face identification \cite{koch_2015}, processing sets of elements \cite{Maron2020OnLS,Zaheer2017DeepS} and object tracking \cite{Zhang2018StructuredSN,Li2019SiamRPNEO,wang2019fast}. 

\section{Variance-Aware Training}

\subsection{About explicit minimization of the variance error}
In the real-world problem setting, the general framework for training a deep learning model considers the training of $M$ models ($M$-fold cross-validation framework) using different training subsets and measuring the model's performance on validation subsets. After selecting the best-performing model or aggregating trained models as an ensemble, the final model performance is estimated using an unseen testing subset. In this scenario, the expectation of the models' performance on the validation subsets characterizes the \textbf{variance error} of the trained ensemble \cite{Heskes}.

This prompted us to ask: Can we create a neural network where the model's variance error can be minimized directly, as a part of the loss function, while maintaining the integrity of the cross-validation procedure (i.e., keeping the target of the validation data hidden)? In trying to answer this question, we summarized the generic theoretical background of the proposed VAT method in the following theorem: 

\textbf{Theorem 1.} \textit{Let $\bar{P}$ be a likelihood distribution learned by neural network $G$ using a \textbf{finite} set of samples from the training subset. $\hat{P}$ is a likelihood distribution estimated by neural network $G$ using a \textbf{finite} set of samples from the validation subset. $Q$ is a likelihood distribution estimated by neural network $G$ using an \textbf{infinite} set of samples from the testing subset. Neural network $G$ is a compact universal approximate of $Q$ (in the sense of $\|\cdot\|_{\infty}$) if the optimization objective of $G$ is defined as follows:}

\begin{equation} \label{eq:eq1}
\max_{\bar{P}} \left[  \bar{P}(\hat{t}_{tr}=t_{tr} |x,\theta ) - \lambda  \mathbb{E}[D_{KL}(\hat{P}(x_{val},\theta | t_{val}) \| \bar{P}(x_{tr},\theta | t_{tr}))]\right] 
\end{equation}

\textit{, when $x_{tr} \in \mathbb{X}_{tr}, x_{val} \in \mathbb{X}_{val}$ are finite sets of training and validation samples,  $\theta$ is the set of trainable parameters of $G$, $t_{tr}$ is the ground true labels, $\hat{t}_{tr}$ is the model predictions, $D_{KL}$ is the Kullback–Leibler divergence between $\hat{P}$ and $\bar{P}$, and $\lambda$ is the mixing coefficient between the two optimization objectives.}

Utilizing the Bias-Variance Trade-off decomposition, we formulated the optimization objective, which focused on minimizing the model's variance error as a minimization of the expectation of the Kullback–Leibler divergence ($D_{KL}$) between likelihood features distributions from the training and validation subsets. Further, the neural network $G$ is supposed to be a universal approximation of $Q$, therefore, $G$ should have enough explicit power to express feature representations in $\bar{P}$. The full proof of the Theorem 1 is available in the Supplementary material (Section 8.1). 

The proof of Theorem 1 presents several intuitive observations. Firstly, we demonstrate that minimizing the $D_{KL}$ between distribution of classes from training and validation subsets is equivalent to explicit minimization of the model’s variance error. Moreover, according to Bayes' rule, if the data present in training and validation subsets share the same inherent posterior probability distributions (i.e., have matching feature-target relations), the minimization of the model's variance error can be reached by the matching of learned likelihood distributions (feature activation map distributions in the case of CNN networks). Therefore, prior knowledge of targets is not required, which maintains the cross-validation integrity. 

Secondly, since the optimization objective (1) is expressing the min-max game (the first term is tasked with maximizing the predictive capability and the second term is tasked with minimizing a feature distribution gap), $D_{KL}$ characterizes the explicit regularization of $G$, parameterized by the mixing coefficient $\lambda$. Thus, introducing the variance into the cost function does not decrease the degree of freedom of the model's parameter space and simultaneously leads the model to find an optimal min-max equilibrium of the bias and variance error components in Equation 5. Accordingly, we explored the relationship between the parameter $\lambda$ and the resulting model's error and provided experimental results in Section 4.

Lastly, we demonstrated that the bias error component can only be minimized implicitly. If the morphological diversity of $\bar{P}$, due to limited available data, is not sufficient to express the diversity of the feature representations in $Q$ (out-of distribution generalization problem), the bias component becomes the prevailing component in the BVTD, which is a limitation for the proposed method.

\subsection{Model Architecture}
The generic Self-Supervised Learning framework is constrained by the problem setting where only a limited portion of the annotated images ($x_{tr} \in X_{tr}, x_{val} \in X_{val}$) is available for training and a relatively larger set of unlabeled images is available for pre-training $x_{Pre} \in X_{Pre}$. Although we observed that our method could operate using only the validation subset $X_{val}$ (not requiring the use of a $X_{Pre}$ subset), the implementation of VAT described below does include sampling from the pre-training subset $X_{Pre}$ to ensure a fair comparison with other state-of-the-art methods. We also assumed that $X_{Pre}$ includes more data samples and samples of more diverse spatial morphology, compared to $X_{val}$, which resulted in improved generalization.

\begin{figure*}[!t]
\centerline{\includegraphics[width=\linewidth]{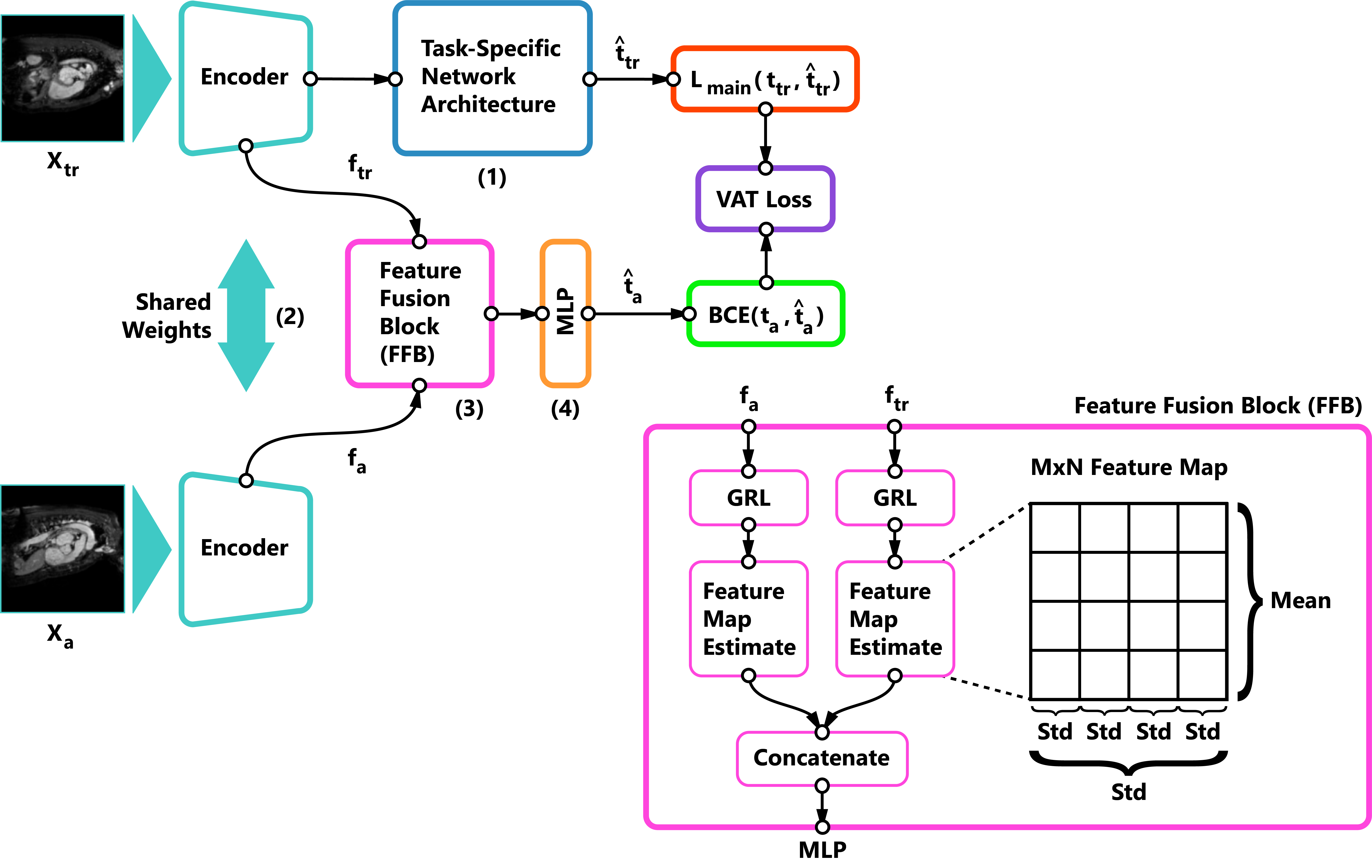}}
\caption{Generic structure of Variance-Aware Training framework. MLP (4) is tasked to classify if 2
images are part of the same subset (training or validation/pre-training). Due to incorporated GRL layers, all gradients that lead to distinguishing subsets will be reversed, which leads to forgetting subset-specific information, i.e., explicit minimization of the model variance. After training, only main network (1) will be used for inference.}
\label{agg}
\end{figure*}

Following Equation 1, our method is constructed using four essential components (Figure 1). The generic CNN model $S: \{e(\cdot),d(\cdot)\}$ (1) is tasked with an arbitrary learning objective $L_{main}: \{d(e(x_t))=\hat{t}_{tr} \to t_{tr}\}$, where $d(\cdot)$ represents the task-specific network architecture (decoder for segmentation tasks, projection network for classification, etc.), and $t_{tr}$ and $\hat{t}_{tr}$ are arbitrary ground truth labels and model predictions, respectively. A SNN (2) is constructed from the encoder $e(\cdot):\{e_1(\cdot),...e_k(\cdot)\}$ with $k \in K$ sequential convolution blocks, which generates feature representations  $f_{tr}:\{e_1(x_{tr}),...e_k(x_{tr})\}$ and $f_{a}:\{e_1(x_{a}),...e_k(x_{a})\}$ where $x_{a}$ is a randomly sampled image from the training or pre-training subset (Algorithm 1) and $x_{tr}$ is an image sampled from the training subset.  An Adversarial Multi-Layer Perceptron (MLP) network $D$ (4), which is constrained by a Binary Cross-Entropy cost function, is tasked with classifying the target $t_{a}:\{t_{a}=0 | x_{tr},x_{a}\in X_{tr}; \: t_{a}=1 | x_{tr} \in X_{tr}, x_{a} \in X_{Pre} \}$ as a special case of $D_{KL}(\bar{P} \| \hat{P})$ in Equation 1. Lastly, the Feature Fusion Block (FFB) (3) estimates statistics of the feature representations $f_{tr}$ and $f_{a}$ (Figure 1). It incorporates the Gradient Reversal  Layer \cite{uada} (GRL) such that all gradients passed between $S$ and $D$ are reversed, meaning that any improvement to the classification performance of $D$ leads to “forgetting” associated feature representations obtained from the encoder $S$. The cost function of the multi-task learning CNN $G:\{S,D\}$ is expressed in Equation 2.

\begin{equation} \label{eq:loss}
L_{total} = L_{main}(t_{tr},\hat t_{tr}) + \lambda BCE(t_{a},\hat t_{a})
\end{equation}

It is important to note the principal differences between the described architecture and the theoretical justification provided in Section 3.1. Firstly, to reduce the computational complexity, the proposed implementation considers optimizing only one model instead of $M$ models in one graph (Equation 2). Despite the fact that such simplification strictly violates the criteria of variance error minimization, we observed a high efficiency of the proposed method, supported by empirical observations of the model's variance error reduction, described in Section 4.2. 

Lastly, we parameterized the $D_{KL}$ with adversarial network $D$, which resulted in improving the stability of convergence. We also introduced structural changes into the adversarial domain adaptation legacy methods. \cite{uada,kim_2019} describes a model architecture where the adversarial network is connected to the last feature embedding layer that conveys the global feature representations $e_K(x)$. At the same time, penalizing only global feature representations causes an optimization issue for the VAT method because the information flow tends to avoid introduced adversarial constraints on the loss surface. To explore this effect, we propose two types of fusion: early aggregation where feature maps $(f_{tr},f_{a})$ from all CNN blocks are connected to the FFB, and late aggregation $(e_K(x_{i}),e_K(x_{a}))$, which is similar to previous studies \cite{uada,kim_2019} (Figure 2). The early aggregation method with $K$ CNN block requires $K$ Local Reparameterization Tricks \cite{Blum2015VariationalDA} to backpropagate through a random node. This could be computationally expensive, and therefore, we replaced it with parametric statistical estimates FFB (Figure 1) to overcome this issue.

\begin{figure*}[!t]
\centerline{\includegraphics[width= 11 cm]{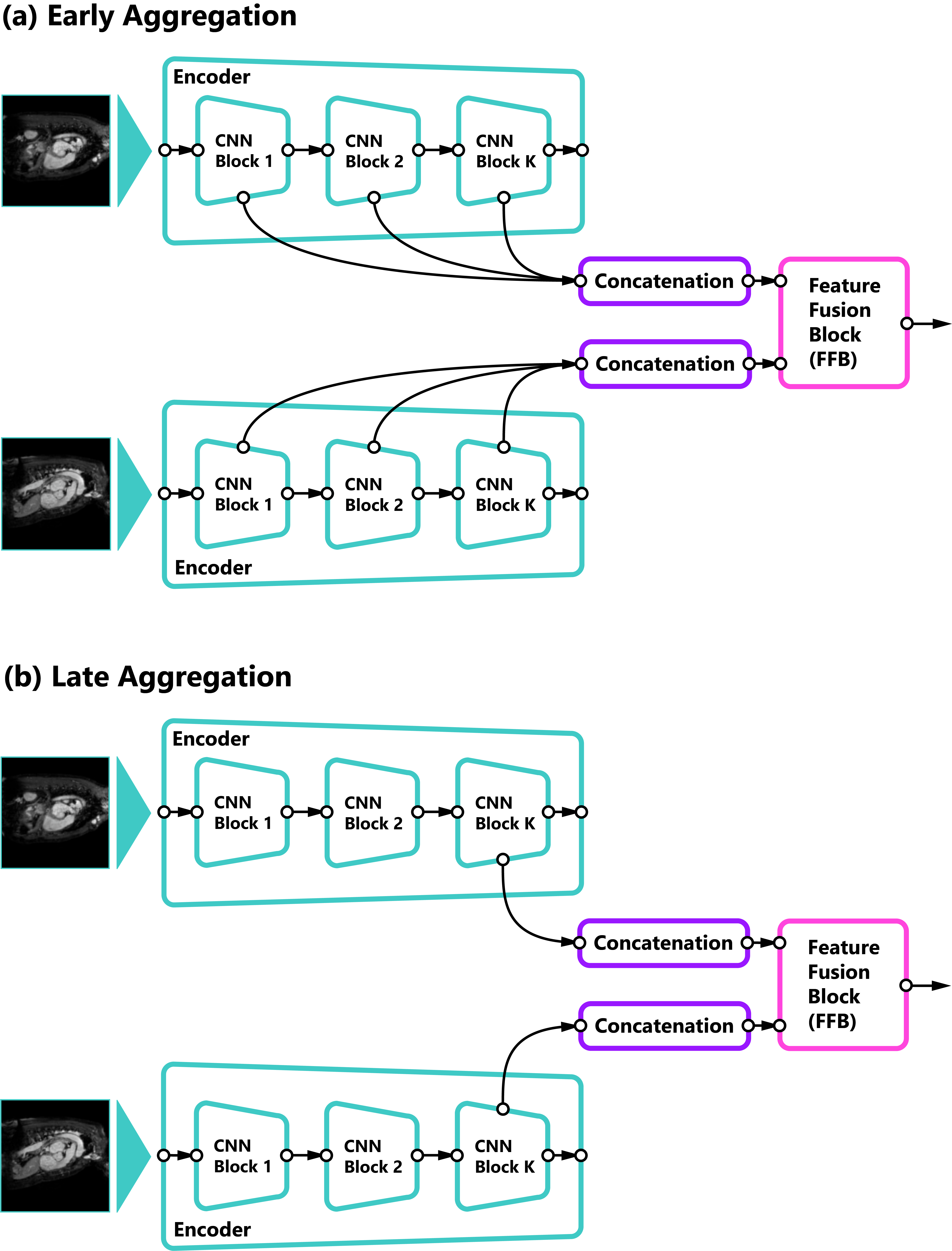}}
\caption{Proposed alternatives for aggregation methods: (a) early aggregation when all layers from the encoder are penalized by VAT, (b) late aggregation when only global feature representations are penalized.}
\label{agg}
\end{figure*}

Only the main network $S$ is used for final inference after VAT, which will not increase the computational and space requirements of the resulting base neural network.

\centerline{\begin{minipage}{8cm}
\begin{algorithm}[H]

\KwResult{$x_{a},t_{a}$}
    $x_{tr} \in X_{tr}$ \\
    $X_{pre}$ \\
    $a \gets Uniform[0,1]$ \\

\textbf{Instructions:}\\
  \eIf{$a \leq 0.5$}{
    $x_{a} \in X_{tr}, \: x_{a}\ne x_{tr}$\\
    $t_{a}=1$\\
   }{
   $x_{a} \in X_{pre}$\\
   $t_{a}=0$\\
  }
 \caption{Sampling of the auxiliary image $x_{a}$}

\end{algorithm}

\end{minipage}
}

\section{Experiments}

\begin{table}[!t]
\centering
\caption{Summarised description of experimental datasets.}
\begin{tabular}{|c||c|c|c|}
\hline
Characteristic & ACDC & APTOS  & PCam\\
\hline
\hline

\hline
Domain  & MRI images & Fundus photography & Histopathologic scans  \\
\hline

Learning objective  & Semantic segmentation & Ordinary regression & Binary classification \\
\hline

Color map & Grayscale & RGB & RGB \\  
\hline

Spatial correlation   & Yes (3D scans)	& No & No  \\
between samples & & & \\

\hline
 & Healthy, previous  & & \\
 & myocardial infarction, & Healthy, mild, &  Metastatic and \\
Population  & dilated cardiomyopathy, & moderate, severe, & normal tissues\\
 & hypertrophic  & proliferative  &\\
 & cardiomyopathy, & retinopaty &\\
 & abnormal right ventricle & & \\

\hline
 &Small and large  &   & \\
& adjacent segments, & Variety of abnormal & Low resolution \\
Morphological features & variability of patterns & tissues &  \\
& between patients & & \\
& populations & & \\
\hline
Number of samples & 100 volumes & 3662 images & 220025 images \\
&  (1902 slices) & &  \\

\hline
 \end{tabular}%
\end{table}

\subsection{Experimental setup}
\textbf {Datasets:} We evaluated the proposed method on 3 diverse tasks using publicly available medical imaging datasets. (1) The ACDC semantic segmentation dataset \cite{acdc} includes 100 volumetric images of 1.5T and 3T MRI scans with three expert-annotated structures: left ventricle, myocardium, and right ventricle. (2) The APTOS 2019 dataset \cite{aptos} includes 3662 fundus photography images for ordinary regression of the diabetic retinopathy progression: healthy condition, mild, moderate, severe, and proliferative. (3) PatchCamelyon (PCam) dataset \cite{Bejnordi2017DiagnosticAO} includes 220025 low-resolution histopathologic scans of lymph node sections for classification of healthy and metastatic tissues. We summarized the qualitative and quantitative specifications of these datasets in Table 1.

\textbf{Base model:} We used a 6-layer 2D Dense U-Net \cite{Ding2019DeepRD} constrained by the smoothed Dice macro loss function as a base architecture for semantic segmentation experiments (ACDC). The EfficientNet-b3 \cite{Tan2019EfficientNetRM} model, constrained by the mean absolute error and Binary Cross-Entropy loss functions, was used for ordinary regression (APTOS) and classification (PCam) experiments, respectively. EfficientNet-b3 was pre-trained on the ImageNet dataset. All of our models were trained using the Adam optimizer \cite{Kingma2015AdamAM} with a learning rate of 0.001 and batch size of 32. For SSL tasks, we optimized the model for 1000 epochs. For downstream SSL tasks, fully supervised and VAT, we optimized models for 3000 epochs. 

\textbf{Pre-processing:} Similar to \cite{Tymchenko2020DeepLA}, we applied a circular crop to APTOS images and selected only the eye zone to prevent over-fitting on image borders. For the APTOS and PCam datasets, before feeding the image into the model, we applied normalization using ImageNet coefficients (mean: 0.485, 0.456, 0.406; standard deviation: 0.229, 0.224, 0.225). For the ACDC dataset, we applied channel-wise standard scaling to the images. Images from the ACDC and APTOS datasets were reshaped to resolutions of 154x154 and 256x256, respectively. The optimal set of rigid and non-rigid augmentations were manually determined for all experiments. For the ACDC dataset, we used horizontal flip, random rotation, elastic transform, random crop, and random gamma augmentations. For the APTOS and PCam datasets, we used horizontal and vertical flip, random rotation, random crop and gamma.

\textbf{Comparison of methods:} We compared VAT with state-of-the-art Self-supervised methods including SimCLR \cite{hinton}, Context Prediction (CP) \cite{Doersch2015UnsupervisedVR}, and Rotation \cite{Gidaris2018UnsupervisedRL}. We also provided an upper-boundary (UB) fully-supervised benchmark, which emulated the setup when all data is available for training. For all proposed methods, we reported the best performance evaluated on the unseen testing subsets $X_{test}$ obtained after optimizing the mixing coefficient $\lambda$ on the validation subset.

\begin{figure*}[b!]
\centerline{\includegraphics[width=12cm]{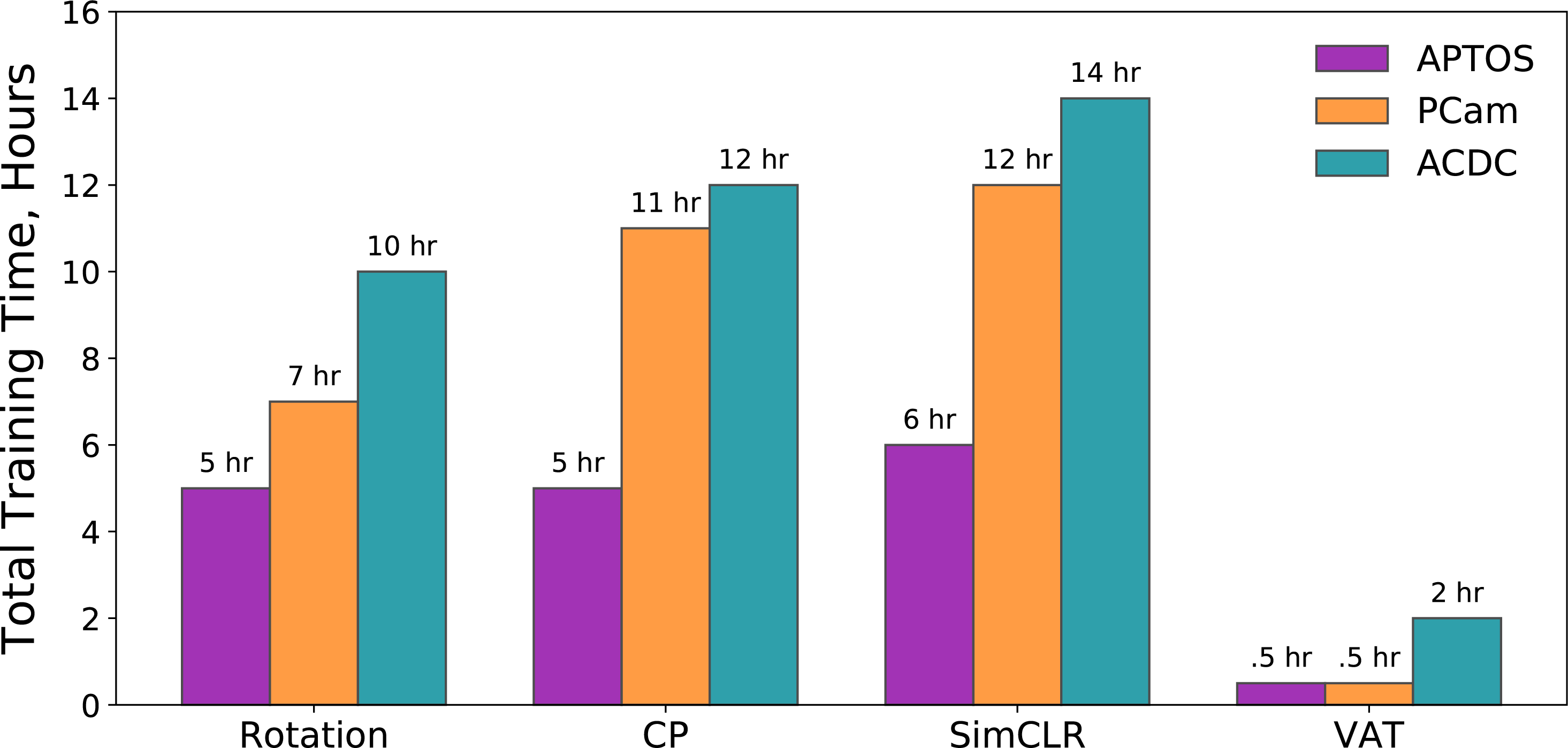}}
\caption{Comparison of the total training time of VAT method with self-supervised methods (pre-train and downstream tasks) for ACDC $X_{tr}=8$, APTOS $X_{tr}=234$, and PCam $X_{tr}=83$  setups.}
\label{agg}
\end{figure*}

\textbf{Cross-validation strategy:} For the ACDC dataset, we split the original data into $X_{pre}$ (80 volumes) and $X_{test}$ (20 volumes) subsets. For the limited annotation problem setups, we selected $X_{tr}=2$, $X_{tr}=4$, and $X_{tr}=8$ 3D volumes from $X_{pre}$ for model training, leaving 2 volumes for validation ($X_{val}=2$). The rest of $X_{pre}$ images, not including  $ X_{tr}$ and $X_{val}$ subsets, were considered as images without labels and were available only for pre-training. For the Upper Bound (UB) setup, we split $X_{pre}$ into $X_{tr}=60$ and $X_{val}=20$ volumetric images. We applied the same splitting strategy for the APTOS experiment with $X_{test}=733$, $X_{tr}=59$, $X_{tr}=122$, $X_{tr}=234$, $X_{pre}=2811$ and $X_{val}=29$ images for the limited annotation problem setups, and $X_{tr}=2343$, $X_{val}=293$ and $X_{test}=733$ for the UB setup. Finally, we split the PCam dataset into $X_{test}=44005$, $X_{tr}=22$, $X_{tr}=44$, $X_{tr}=83$, $X_{pre}=175975$ and $X_{val}=23$ images for the limited annotation experiments and $X_{tr}=140816$, $X_{val}=3521$ and $X_{test}=44005$ for the UB experiments. For all experiments, we trained the model using $X_{train}$, using $X_{val}$ for early stopping, and performed the final evaluation on the unseen $X_{test}$ subsets. The summarized schema of cross-validation split is reflected in Figure 5 (Section 8.2).

\textbf{Evaluation:} Dice macro similarity (without background), Area Under the Receiver Operating Characteristic Curve (AUC-ROC), and quadratic weighted Kappa scores were used to evaluate the segmentation (ACDC), classification (PCam), and ordinary regression (APTOS) performance, respectively. For all experiments, we report the mean scores evaluated on the $X_{test}$ subset over 5 runs. 

\subsection{Experimental results}
A comparison of the proposed method (VAT) with state-of-the-art self-supervised methods is presented in Tables 2-4. VAT with late aggregation did not succeed in most experiments while the early aggregation technique produced strong results, outperforming the baseline models. As discussed above, we hypothesized that late aggregation would not provide explicit enough penalization of the feature representations, which would lead to the de-synchronization of local and global feature representations during training and therefore, degrade the model's performance.

\begin{table}[!h]
\centering
\caption{Comparison of the proposed method with pre-training methods,  ACDC dataset.}
\begin{tabular}{|c||c|c|c|c|}
\hline
Method & $X_{tr}=2$ & $X_{tr}=4$  & $X_{tr}=8$ & $X_{tr}=60$ (UB)\\
\hline
\hline
Baseline & 0.700 & 0.766 & \textbf{0.834} & 0.902 \\

SimCLR & 0.721 & 0.748 & 0.796 & - \\

CP & 0.743	& 0.790 & 0.799 & - \\

Rotation  & 0.744 & 0.791 & \textbf{0.834} & - \\
\hline

VAT, early agg. (ours) & \textbf{0.749} & \textbf{0.814} & \textbf{0.834} & - \\

VAT, late agg.(ours) & 0.616 & \textbf{0.795} & 0.546 & - \\
\hline
 \end{tabular}%
\end{table}

\begin{table}[!h]
\centering
\caption{Comparison of the proposed method with pre-training methods,  APTOS dataset.}
\begin{tabular}{|c||c|c|c|c|}
\hline
Method & $X_{tr}=59$ & $X_{tr}=122$  & $X_{tr}=234$ & $X_{tr}=2343$ (UB)\\
\hline
\hline
Baseline & 0.804 & 0.851 & 0.852 & 0.909 \\

SimCLR & 0.619 & 0.769 & 0.789 & - \\

CP & 0.822 & 0.847 & 0.849 & - \\

Rotation & 0.827 & 0.855 & 0.863 & - \\
\hline

VAT, early agg. (ours) & \textbf{0.851} & \textbf{0.863} & \textbf{0.868} & - \\

VAT, late agg.(ours) & 0.452 & 0.455 & 0.462 & - \\
\hline
 \end{tabular}%
\end{table}

\begin{table}[!h]
\centering
\caption{Comparison of the proposed method with pre-training methods, PCam dataset.}
\begin{tabular}{|c||c|c|c|c|}
\hline
Method & $X_{tr}=22$ & $X_{tr}=44$  & $X_{tr}=83$ & $X_{tr}=140816$ (UB)\\
\hline
\hline
Baseline & 0.776 & 0.745 & 0.726 & 0.989 \\

SimCLR & 0.781 & \textbf{0.817} & \textbf{0.860} & - \\

CP & 0.706 & 0.794 & 0.814 & - \\

Rotation & 0.739 & 0.760 & 0.824 & - \\
\hline

VAT, early agg. (ours) & \textbf{0.806} & 0.809 & 0.833 & - \\

VAT, late agg.(ours) & 0.796 & 0.777 & 0.822 & - \\
\hline
 \end{tabular}%
\end{table}

Compared to the SSL methods, VAT either matched or surpassed the top SSL scores. SSL methods resulted in a range of performances on different tasks. SimCLR provided the best performance for the PCam experiment but failed on others resulting in a performance that was worse than the baseline model. Rotation and CP methods produced a strong performance on all tasks, improving the baseline, but VAT consistently matched or outperformed those methods. In total training GPU time, SSL pre-training and fine-tuning took more than 10 hours of GPU time on 2 RTX 2080Ti units for all experiments, while VAT demonstrated almost the same training time as the baseline model, which was only 2 hours for ACDC and 0.5 hours for APTOS and PCam experiments (Figure 4). 

Additionally, we faced the challenge of finding an optimal combination of hyper-parameters and augmentations for SSL pre-training tasks. In contrast, our method required only selecting the mixing coefficient $\lambda$ and used the same augmentations and model architecture as the baseline solution. We suggest that combined with fast convergence, VAT could drastically decrease the time spent on model research. 
We also empirically demonstrated that VAT does indeed minimize the model's variance. The positive effect of introducing VAT into the baseline model decreases with the increase in the dataset size (Tables 2-3) and we found the same correlation with respect to the optimal mixing coefficient $\lambda$ (Figures 4a-4f). In the general case for a model with a fixed capacity, the variance error decreases with an increase in the dataset size and therefore, it reduces the variance error of the model.

Furthermore, we found the PCam classification experiment resulted in the most expressive VAT performance. When introducing more training examples, the AUC-ROC score degraded from 0.776 to 0.726 due to an increase in unrepresentable samples, which lead to over-fitting. In contrast, our method penalizes shifted feature representations, bringing back the positive correlation between the model's performance and dataset size. Moreover, increasing the number of unrepresentable training samples increased the optimal $\lambda$ (Figures 4g-4i), which supports our observations.

\section{Limitations}
The practical application of the proposed method is limited by the fundamental assumption of the morphological compactness of the training data. We assume that the approach will provide a poor performance when applied to computer vision problems with complex environmental setup (light, position of camera, diverse set of objects and camera pose) where the bias error will be the dominant component of the BVTD. Additionally, since the proposed method requires storing twice as many gradients for the encoder, it increases GPU memory consumption compared to the baseline model, which could be a limitation for tasks requiring large models (for example, object detection) or large batch size.

\begin{figure*}[h!]
\centerline{\includegraphics[width=\linewidth]{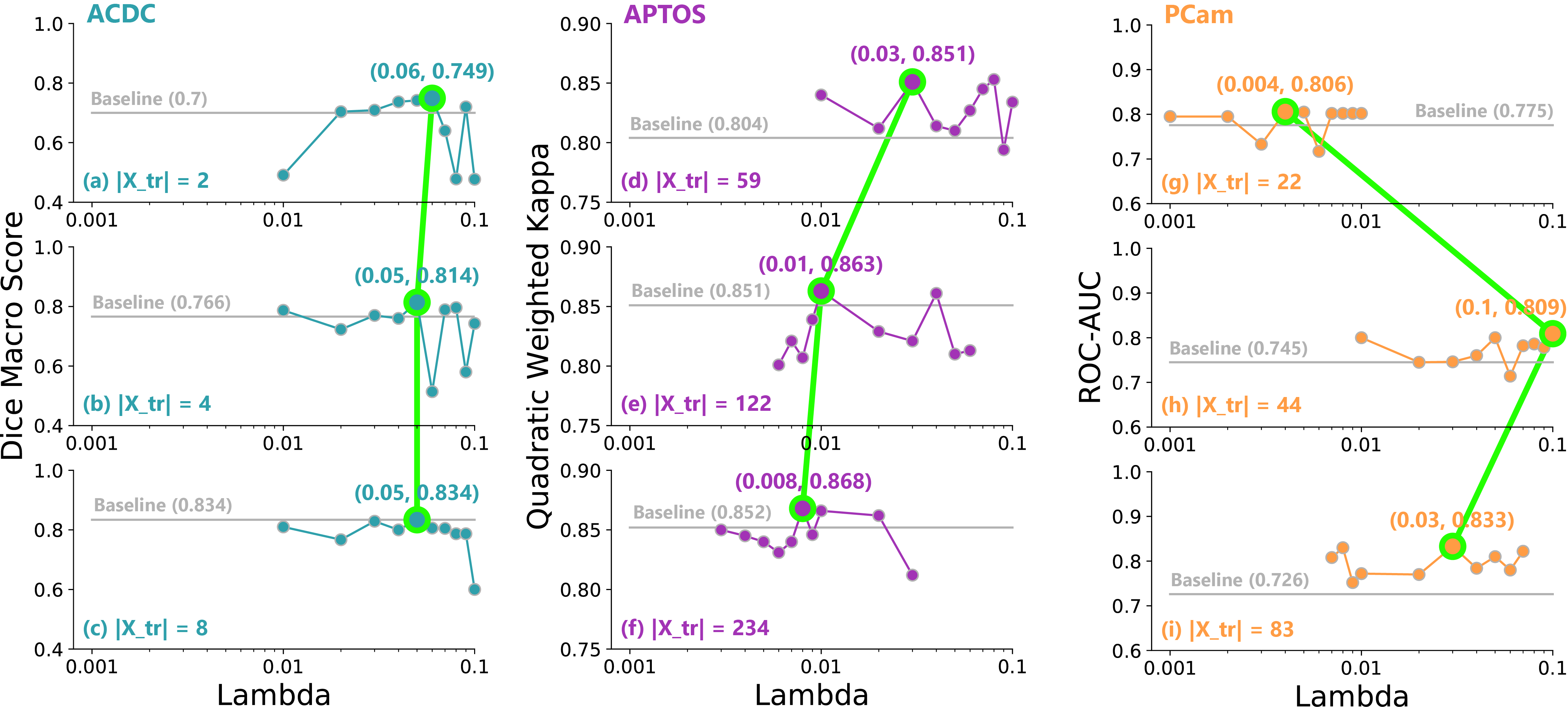}}
\caption{Exploration of the relationships between selected mixing coefficient $\lambda$ and final model performance: (a), (b), (c) ACDC experiments; (d), (e), (f) APTOS experiments; (g), (h), (i) PCam experiments.}
\label{agg}
\end{figure*}

\section{Conclusion}
In this paper, we proposed a novel method for medical imaging model training, called Variance-Aware Training, which focuses on the direct optimization of model estimation variance. Our contributions include both a theoretical formulation and an experimental analysis of the proposed method. As a result, we demonstrated that, compared to Self-Supervised methods, our approach achieved the same or superior performance without the need for pre-training, which resulted in a 4-11\% improvement on various medical imaging tasks. In addition to these performance improvements, our method was able to significantly reduce the training time and simplify model development.

\section{Acknowledgments}
We would like to thank Animesh Garg for fruitful discussions and feedback on the draft. Also, we would like to thank Laussen Labs and SickKids Hospital for their support.

\printbibliography

\newpage

\section{Supplementary Material}
\subsection{Proof of the Theorem 1}
According to the Universal Approximation Theorem \cite{Cybenko1989ApproximationBS,Lu2020AUA}:

\begin{equation} \label{eq:eq3}
\exists G : \sup_{x \in R^{n}}| \mathbb{E}[f(x)] - \mathbb{E}[G(x,\theta)]  | < \epsilon, \epsilon \to 0 \quad \forall  f: R^{n}\to R^{m} \quad \forall x \in
R^{n}
\end{equation}

where $f$ is an arbitrary continuous polynomial function. In general case, the optimization objective is represented as fitting a probabilistic distributions, and the Equation 1 could be reformulated as:

\begin{equation} \label{eq:eq4}
\exists G : \sup_{x \in R^{n}} | Q - \bar{P}  |< \epsilon, \epsilon \to 0 \quad 
\end{equation}

, when $\bar{P}$ represents a probabilistic distribution of finite set of training samples, and $Q$ represents probabilistic distribution of infinite set of testing samples (general population).

By definition, Equation 4 expresses the Total Distance of Probability Measures (TDPM) between $\bar{P}$ and  $Q$. Following Pinsker's inequality, the TDPM is upper bounded by the Kullback–Leibler divergence ($D_{KL}$) of the estimated probability distributions (Equation 5). Thereby, minimization of the KL divergence leads to minimization of the TDPM between the $f$ and it's approximation $G$ (Equation 6,7).

\begin{equation} \label{eq:eq5}
\sup_{x \in R^{n}} | Q - \bar{P} | \leq \sqrt{\frac{1}{2} D_{KL}(\:  Q \: \| \bar{P}  \:)}
\end{equation}

\begin{equation} \label{eq:eq6}
 min_{x \in R^{n}}  |   \mathbb{E}[f(x)] -  \mathbb{E}[G(x,\theta)] \ | \iff  min_{x \in R^{n}}D_{KL}(\: \mathbb{E}[f(x)]     \: \| \:  \mathbb{E}[G(x,\theta)] \:)
\end{equation}

\begin{equation} \label{eq:eq7}
 min_{x \in R^{n}}  | Q - \bar{P} | \iff  min_{x \in R^{n}}D_{KL}(\: Q   \: \| \bar{P} \:)
\end{equation}

The expectation of the error $\mathbb{E}[D_{KL}(Q \| \bar{P})]$ is characterised by the BVTD \cite{Heskes} (Equation 5), where $\hat{P}$ represents the distribution estimated by $G$ on the validation subset, which is also specified as an average distribution $\hat{P} = \frac{1}{2}(\bar{P} + Q)$). 

\begin{equation} \label{eq:eq8}
\mathbb{E}\left[ D_{KL}(Q \| \bar{P}) \right] =  \underbrace{D_{KL}(Q \| \hat{P})}_{Bias \:error} + \underbrace{\mathbb{E}[D_{KL}(\hat{P} \| \bar{P})]}_{Variance \:error} + \sum_{}^{N}\underbrace{Qlog(Q)}_{Bayes \:error}
\end{equation}

According to Bayes’ rule:

\begin{equation} \label{eq:eq9}
\hat{P} = \hat{P}(t_{tr} | x_{tr},\theta) = \frac{\hat{P}(x_{tr},\theta | t_{tr})\hat{P}(t_{tr}) }{\hat{P}(x_{tr},\theta)}
\end{equation}

\begin{equation} \label{eq:eq10}
\bar{P} = \bar{P}(t_{val} | x_{val},\theta) = \frac{\bar{P}(x_{val},\theta | t_{val})\bar{P}(t_{val}) }{\bar{P}(x_{val},\theta)}
\end{equation}

, when $x_{tr},x_{val}$ - denotes the input of the DNN, $t_{tr},t_{val}$ - denotes the ground-truth output of DNN, $\theta$ - denotes parameters of DNN. In this case, if prior and evidence distributions are equal in training and validation subsets, the resulting variance error is characterised as:

\begin{equation} \label{eq:eq11}
Error_{variance} = \mathbb{E}[D_{KL}(\hat{P} \| \bar{P})] \sim  \mathbb{E}[D_{KL}(\hat{P}(x_{val},\theta | t_{val}) \| \bar{P}(x_{tr},\theta | t_{tr}))]
\end{equation}

Therefore, by introducing $ \mathbb{E}[D_{KL}(\hat{P}(x_{val},\theta | y) \| \bar{P}(x_{val},\theta | y))] $ as an additional constraint into the objective function, the variance term can be optimized \textit{explicitly}. After applying Lagrangian relaxation, the resulting multi-task objective of the DNN is characterized by Equation 1. $\square$

\subsection{Cross-validation strategy}

\begin{figure*}[h]
\centerline{\includegraphics[width=\linewidth]{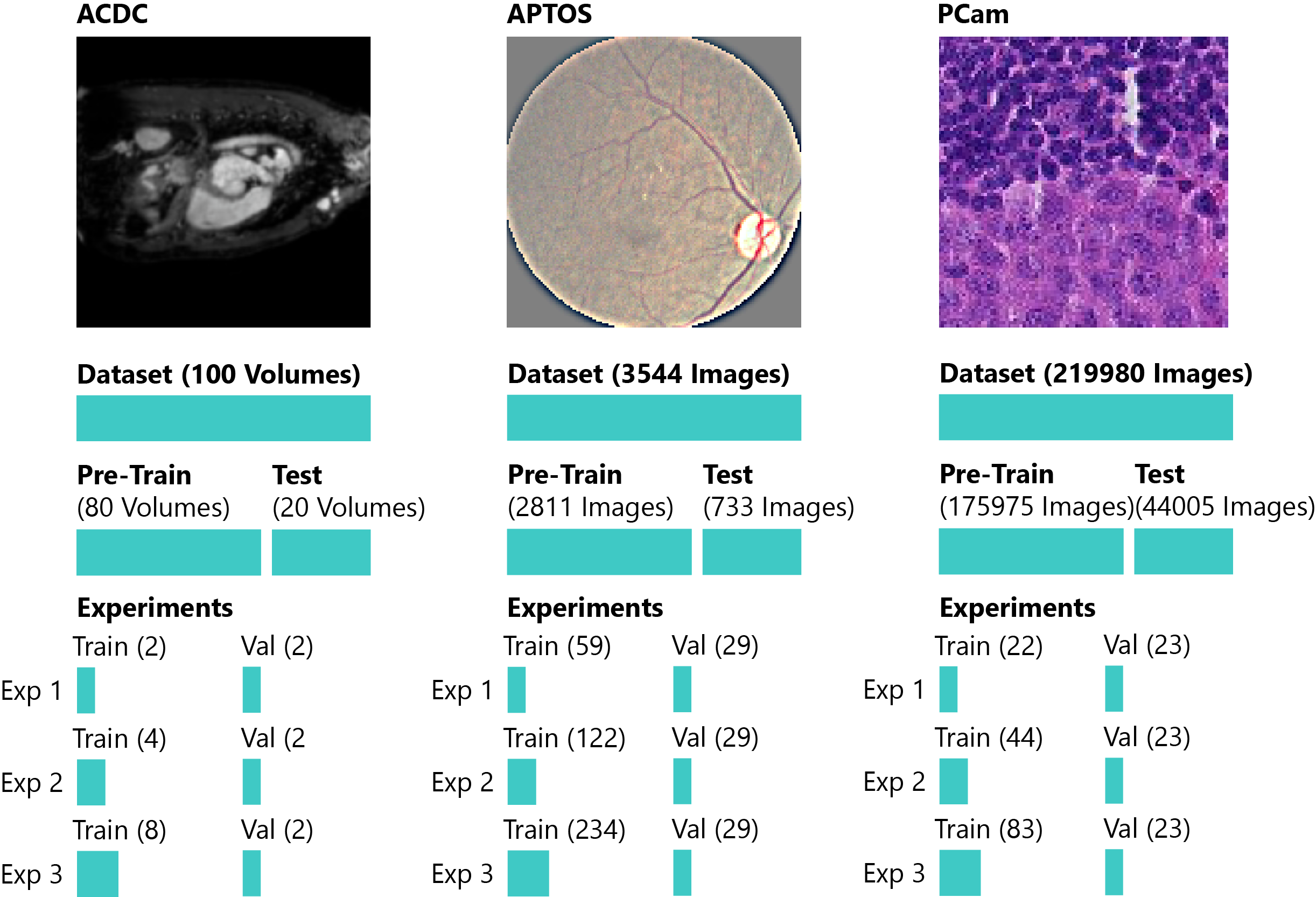}}
\caption{Visualization of summarized cross-validation strategy for all experimental setups.}
\label{cv}
\end{figure*}

\end{document}